\documentclass[letterpaper]{article} 
\usepackage{aaai2026}  
\usepackage{times}  
\usepackage{helvet}  
\usepackage{courier}  
\usepackage[hyphens]{url}  
\usepackage{graphicx} 
\usepackage{natbib}  
\usepackage{caption} 
\usepackage{algorithm}
\usepackage{algorithmic}

\usepackage{newfloat}
\usepackage{listings}
\DeclareCaptionStyle{ruled}{labelfont=normalfont,labelsep=colon,strut=off} 
\floatstyle{ruled}
\newfloat{listing}{tb}{lst}{}
\floatname{listing}{Listing}
\usepackage{tabularx}
\usepackage{booktabs}
\usepackage{amsmath}
\usepackage[capitalize]{cleveref}
\usepackage{subcaption}
\usepackage{tcolorbox}
\newtcolorbox{promptbox}[2][]{
  colback=gray!5!white,       
  colframe=gray!20!black,     
  title=#2,
  sharp corners,              
  boxrule=0.8pt,              
  left=2mm, right=2mm, top=1mm, bottom=1mm, 
  #1                         
}

\newif\ifisextended
\isextendedtrue      

\newcommand{\appref}[1]{%
  \ifisextended
    \cref{#1}%
  \else
    the extended version%
  \fi
}

\ifisextended
    \nocopyright
\fi

\title{Text-to-Scene with Large Reasoning Models}
\author{
    Frédéric Berdoz,
    Luca A. Lanzendörfer,
    Nick Tuninga,
    Roger Wattenhofer
}
\affiliations{
    ETH Zurich\\
    \{fberdoz, lanzendoerfer, ntuninga, wattenhofer\}@ethz.ch
}

\begin{document}

\maketitle

\ifisextended
    \insert\footins{\noindent\footnotesize This is the extended version of the corresponding paper published in the AAAI 2026 proceedings.}
\fi

\begin{abstract}
Prompt-driven scene synthesis allows users to generate complete 3D environments from textual descriptions. Current text-to-scene methods often struggle with complex geometries and object transformations, and tend to show weak adherence to complex instructions. We address these limitations by introducing Reason-3D, a text-to-scene model powered by large reasoning models (LRMs). Reason-3D integrates object retrieval using captions covering physical, functional, and contextual attributes. Reason-3D then places the selected objects based on implicit and explicit layout constraints, and refines their positions with collision-aware spatial reasoning. Evaluated on instructions ranging from simple to complex indoor configurations, Reason-3D significantly outperforms previous methods in human-rated visual fidelity, adherence to constraints, and asset retrieval quality. Beyond its contribution to the field of text-to-scene generation, our work showcases the advanced spatial reasoning abilities of modern LRMs. Additionally, we release the codebase to further the research in object retrieval and placement with LRMs.
\end{abstract}

\begin{links}
    \link{Code}{https://github.com/ETH-DISCO/reason-3d}
    \ifisextended \else
    \link{Extended version}{https://arxiv.org/abs/2509.26091} 
    \fi
\end{links}

\begin{figure*}[t]
  \centering
  \includegraphics[width=\textwidth]{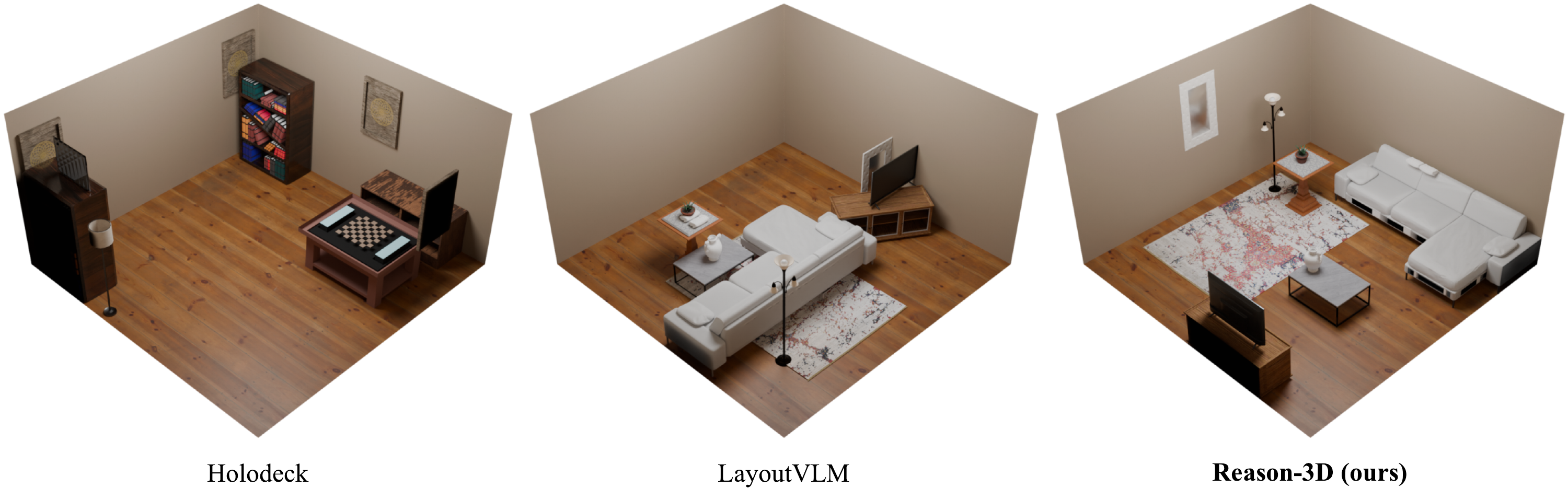}
  \caption{Showcase comparison for object retrieval and placement between Reason-3D and baseline approaches for the instruction \textit{``A cozy living room of size 5 by 5 units. There is a plant on a small table in front of the L-shaped sofa.''}}
  \label{fig:showcase}
\end{figure*}

\section{Introduction}
The demand for customizable 3D scenes is rapidly growing across a wide range of fields, including interior design~\cite{ccelen2024design, fu20213d, dai2017scannet}, video game development~\cite{raistrick2024infinigen, yang2024holodeck, unity_asset_store}, autonomous driving~\cite{gaomagicdrive, mao2025dreamdrive}, robotics~\cite{lee2025dynscene, bu2024closed, po2024compositional, cen2024using}, embodied AI~\cite{yang2024physcene, lee2025dynscene, deitke2022procthor, yang2024holodeck}, and more~\cite{wen20253dscenegenerationsurvey}. Generating plausible and physically‐feasible arrangements of objects from high- to low-level descriptions is therefore at the core of text-to-scene models. Traditionally, scene synthesis has been addressed by specialized models trained on annotated layout datasets or by engineered pipelines that encode spatial priors~\cite{paschalidou2021atiss, tang2024diffuscene}. But these methods tend to constrain the ability of the model in some way, be it context confinement, visual fidelity, physical plausibility, or object transformation freedom. 
Recent advances in training strategies, such as Group Relative Policy Optimization (GRPO), have led to the emergence of Large Reasoning Models (LRMs) ~\cite{shao2024deepseekmath}. Unlike standard LLMs, LRMs are trained to perform multi-step reasoning by leveraging test-time compute through long reasoning traces, enabling them to reason about geometry, context, physical affordances, and object-function relationships~\cite{xu2025towards}. This shift enables context-free scene synthesis, where systems can generate and arrange 3D environments directly from natural language, without relying on predefined templates or training distributions. 
In this work, we present Reason-3D, a modular and highly flexible scene synthesis pipeline that leverages LRMs to construct indoor and outdoor 3D scenes from open-ended textual descriptions. Unlike prior approaches, Reason-3D requires no domain-specific training or architectural constraints. Given a scene prompt, our system extracts relevant objects using a combination of embedding-based retrieval and LLM-based semantic voting across three dimensions: physical, functional, and contextual relevance. We use a dual-stage placement process, where each object is placed in an optimal order autoregressively with the help of an LRM. Finally, Reason-3D refines outputs by making the LRM aware of potential collisions and resolving them. This architecture enables Reason-3D to operate entirely through natural language instructions. It supports complex spatial compositions, unusual object configurations, and even outdoor or hybrid environments without requiring any manual scripting or handcrafted scene rules. 
We evaluate Reason-3D on established indoor scene synthesis tasks where we significantly outperform prior methods in terms of plausibility and structural coherence (see~\cref{fig:showcase}). Furthermore, we highlight the generalization capabilities of Reason-3D by demonstrating applications in outdoor scenes and composition scenarios. Through an ablation study, we find that certain LRMs consistently outperform others across a diverse set of spatial reasoning tasks. 
We summarize our contributions as follows:
\begin{itemize}
    \item We introduce Reason-3D, a novel text-to-scene pipeline that leverages Large Reasoning Models (LRMs) to retrieve and place objects in 3D space, based only on natural language descriptions of the scene, without requiring fine-tuning or pretraining.
    \item Reason-3D significantly outperforms previous baselines by using a dual-phase placement strategy combining autoregressive layout with collision-aware refinement, enabling physically coherent scene synthesis.
    \item Reason-3D generalizes to any scene setting, as demonstrated by our outdoor scene synthesis and open-world compositions. By being able to run out-of-the-box on any object library without requiring scripting or hand-crafted rules.
\end{itemize}

\begin{figure*}[t]
    \centering
    \includegraphics[width=\textwidth]{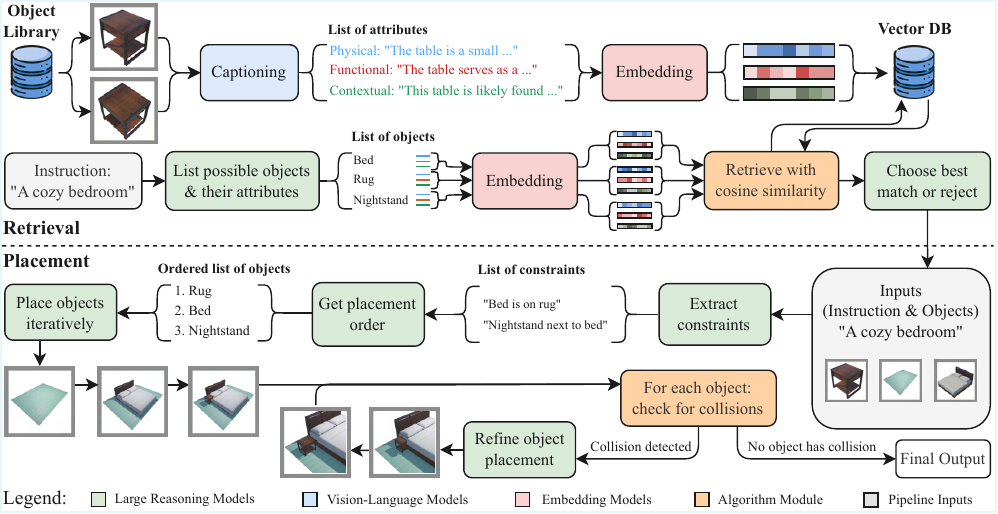}
    \caption{\textbf{Overview of our proposed architecture}. \emph{Retrieval}: We start by processing objects from an asset library, generating images of these objects for captions and orientation. A captioning model creates object descriptions, which are then turned into embedding vectors and stored in a vector database. Given an instruction, we extract a list of objects that would be feasible according to the instruction. and subsequently query the database for such objects. \emph{Placement}: Given the instruction and retrieved objects, we extract a set of constraints to determine an ordered sequence for object placement. Once all objects are placed, the scene is refined by calculating and adjusting for object collisions.}
    \label{fig:overview}
\end{figure*}

\begin{figure*}[t]
  \centering
  \includegraphics[width=\textwidth]{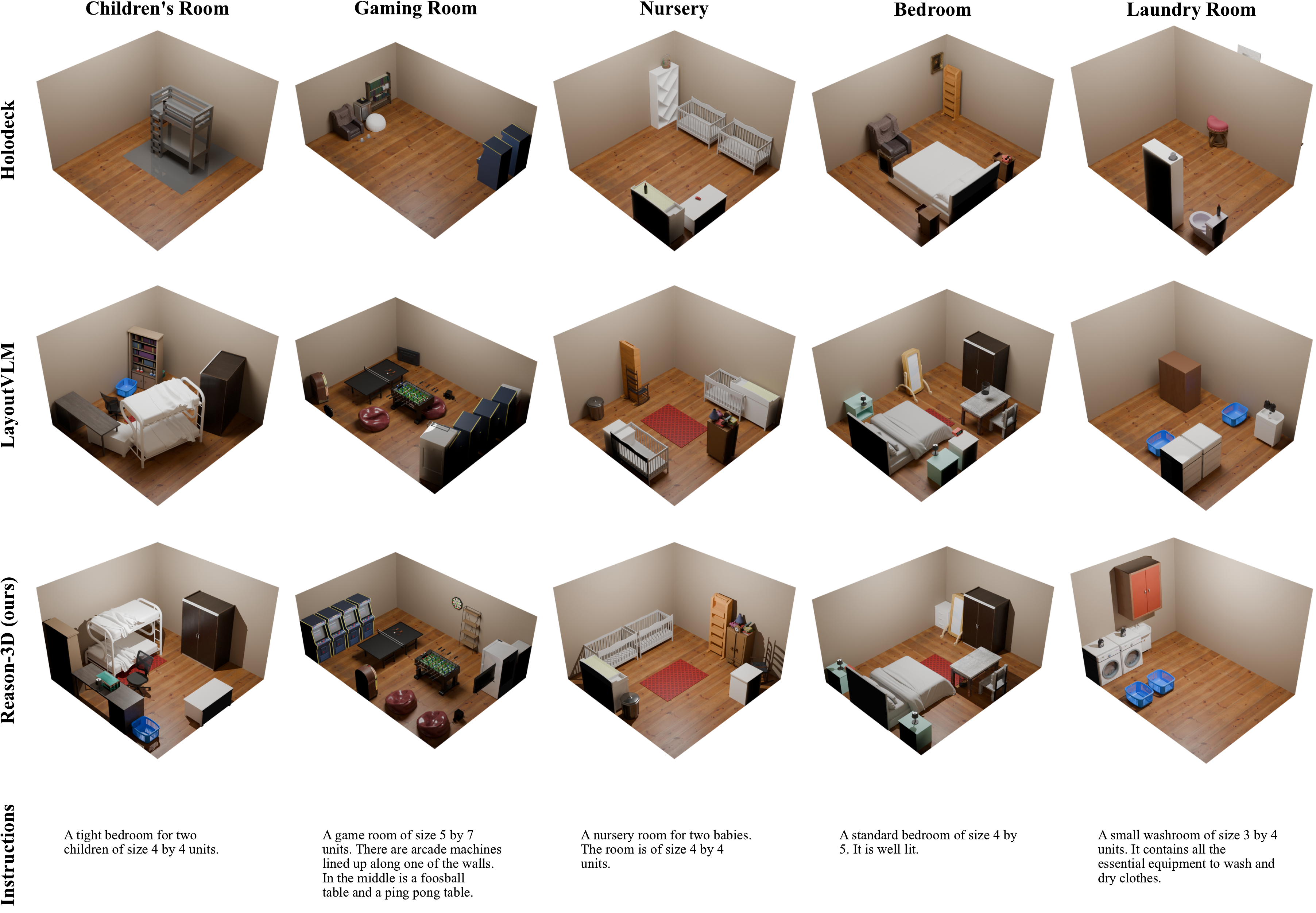}
  \vspace{1em}
  \caption{Qualitative comparison for object retrieval and placement across various scenes. We find that overall, Reason-3D can better follow instructions and place objects reasonably. Compared to Holodeck and Reason-3D, LayoutVLM was not designed to retrieve objects. We use the objects retrieved from Reason-3D for LayoutVLM.}
  \label{fig:extended_comparision}
\end{figure*}

\begin{figure*}[t]
  \centering
  \includegraphics[width=\textwidth]{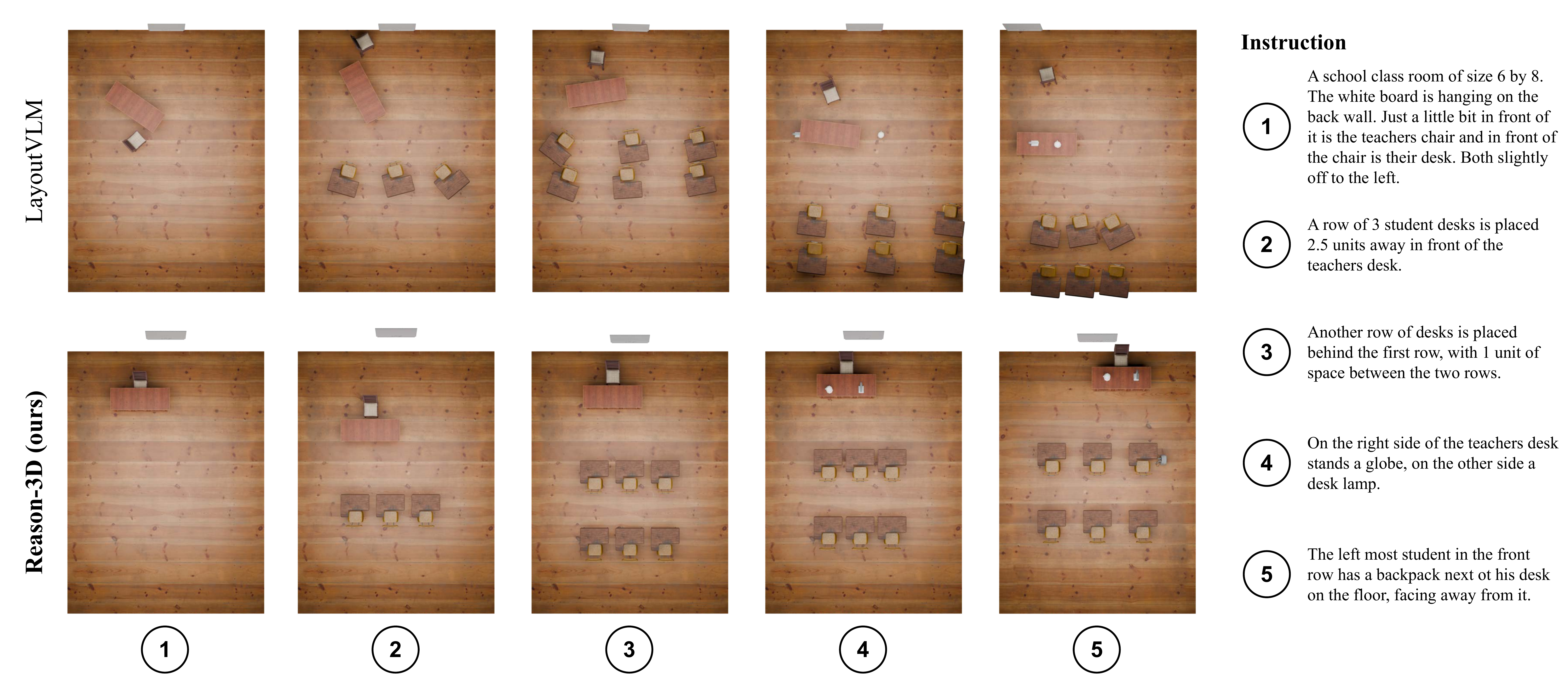}
  \caption{Qualitative comparison of object placement performance when instruction complexity is increased. Every scene is generated from scratch with the entire instruction given up to and including its circled number.}
  \label{fig:evolution_classroom}
\end{figure*}

\section{Related Work}
Recent progress in indoor scene synthesis has primarily followed two distinct approaches. One line of work capitalizes on the strong generative capabilities of image-based models, often employing Neural Radiance Fields (NeRFs) or 3D Gaussian splats as output representations~\cite{schult2024controlroom3d, zhou2024gala3d, epstein2024disentangled, po2024compositional}. While these methods produce visually realistic results, the generated scenes lack object-level separability, making them unsuitable for downstream tasks requiring precise object manipulation or interaction. A second research direction focuses on structured scene generation using intermediate representations, such as scene graphs or layout templates, coupled with curated asset libraries to synthesize 3D environments with discrete objects~\cite{feng2023layoutgpt, yang2024holodeck}.

\paragraph{Neural-3D generation.}
Within this structured generation paradigm, learning-based generative models such as diffusion models have emerged as powerful tools for modeling spatial priors~\cite{hu2024mixed}. DiffuScene~\cite{tang2024diffuscene} applies a denoising diffusion process to generate unordered sets of object attributes, synthesizing layouts from the 3D-FRONT dataset, which comprises 19,000 annotated indoor scenes~\cite{fu20213d}. Other methods use model training with scene graphs~\cite{yang2025mmgdreamer, zhai2024echoscene, lin2023instructscene, zhai2023commonscenes} and without scene graphs~\cite{paschalidou2021atiss, yang2024physcene, ritchie2019fast}. Despite their effectiveness, these models require task-specific training and often struggle to generalize beyond the distributions they were trained on. In contrast, our method performs scene synthesis in a zero-shot setting, relying purely on language-based reasoning without additional fine-tuning or supervision.

\paragraph{Language-based room synthesis.}
The advent of Large Language Models (LLMs) has enabled open-vocabulary 3D scene synthesis, supporting the flexible generation of scenes without dependence on predefined labels or categories. These systems treat the LLM as a ``design assistant'' that outputs a high-level layout or scene graph, which is then materialized into 3D. Holodeck~\cite{yang2024holodeck} and others ~\cite{ccelen2024design, aguina2024open} orchestrate multi-agent LLM systems to generate scene graphs or DSLs (Domain Specific Language), which are later optimized through layout engines. These prompt-based LLM planners are able to interpret open-ended descriptions but often require post-processing or optimization to convert text into numeric layouts, which is an inherent limitation. LayoutVLM~\cite{sun2025layoutvlm} represents the latest advancement in object placement, employing VLMs to generate two mutually reinforcing representations from visually marked images, and a self-consistent decoding process to improve spatial planning.

\paragraph{Direct LLM-to-layout methods.}
A straightforward approach to scene synthesis with LLMs involves directly querying the model for object positions and rotations. LayoutGPT~\cite{feng2023layoutgpt} pioneered this strategy by retrieving relevant example layouts from a database to serve as in-context demonstrations. The GPT-based model then generates a new layout by predicting object bounding boxes conditioned on these exemplars. However, the reliance on a fixed library of example scenes limits generalization, as the diversity of layouts that can be produced is constrained by the database contents. Other approaches fine-tune pretrained LLMs for scene synthesis or editing~\cite{bucher2025respace, yang2024llplace}, but similarly inherit dataset biases that hinder their generalizability. Concurrent work such as DirectLayout~\cite{ran2025direct} prompts an LLM with chain-of-thought reasoning to generate a 2D bird’s-eye view layout, which is then lifted into 3D and refined. While promising, this method is fine-tuned on the 3D-FRONT dataset, again limiting its applicability outside the training distribution. Naively prompting standard LLMs that lack spatial multi-step reasoning capabilities to generate object coordinates often results in physically implausible outputs (e.g., overlapping objects, unrealistic placements, or violations of spatial constraints) because general-purpose non-reasoning LLMs do not inherently model geometry, scale, or collision dynamics. As a result, most recent LLM-based scene synthesis pipelines incorporate auxiliary mechanisms, such as example-based in-context retrieval, spatial reasoning decomposition, or fine-tuned scene grammar models, to compensate for limitations in raw layout prediction and ensure plausible, physically consistent scenes. These hybrid frameworks confirm that LLMs are well-suited for capturing high-level design semantics and following open-ended user instructions, but also underscore the limitations of purely text-based models when tasked with precise spatial generation, in the absence of reasoning steps. In contrast, our method operates entirely with off-the-shelf LLMs exhibiting multi-step spatial reasoning, also referred to as LRMs, requiring no task-specific fine-tuning. This enables generalization to more diverse and open-ended settings, including outdoor environments.

\section{Methodology}
Our proposed method, Reason-3D, consists of multiple stages (see~\cref{fig:overview}). Broadly speaking, the pipeline is divided into two phases: object retrieval and object placement.

\paragraph{Dataset and preprocessing.}
The objects used in our pipeline are sourced from the Objaverse dataset~\cite{objaverseXL}. To ensure reliable operation, these 3D objects must conform to a standardized structure: specifically, each object is required to be upright and oriented consistently along a canonical forward direction. While the system includes runtime mechanisms to detect and correct certain rotational misalignments, overall robustness is significantly improved when operating on a geometrically consistent dataset. Since objects imported from Objaverse do not always adhere to these conventions, we apply a preprocessing step involving a vision-language model (VLM) to analyze the object's appearance and infer its intended orientation. The VLM receives four rendered views of the object and selects the image that most likely represents the front-facing orientation (see~\appref{app:prompts} for the prompting strategy). The identified view is then used to reorient the object accordingly. The remaining discrepancies must be addressed dynamically during layout synthesis by the LRM, provided the object's dimensions and context allow for such adjustments.

\paragraph{Object retrieval.}
For each object in the dataset, we generate a structured textual description using an image captioning model. Each captioning prompt includes two rendered views of the object from different angles and a detailed instruction on what to output (see~\appref{app:prompts} for more details). The resulting structured descriptions are embedded using an embedding model to obtain a high-dimensional representation that supports semantic retrieval.
In parallel, an LRM processes the scene prompt to extract the set of required objects, with each extracted object similarly described using the same tripartite structure. During retrieval, cosine similarity is used to identify the top five closest object candidates based on their embeddings. These candidates are then evaluated by the LRM, which selects the most semantically appropriate instance or determines if no suitable match is available. This ensures that the retrieved instances faithfully reflect the user’s intent, thereby enhancing overall prompt adherence. The retrieved objects are then passed to the placement phase.

\paragraph{Building the scene.}
Object placement and scene construction are carried out in two phases: initial placement and refinement, where Reason-3D leverages the multi-step reasoning of LRMs to sequentially determine each object’s position and rotation in the context of previously placed objects, with full control over all three axes of rotation. The model receives scene constraints, object metadata (name and size), as well as a list of current placements, but no visual feedback (only labeled bounding boxes), requiring it to reason purely from spatial metadata.
Before placement, we address two key challenges. First, implicit spatial relations in the scene prompt are made explicit using an LRM extraction step. This involves querying the LRM to imagine the layout and giving as output object-specific constraints, both relative to other objects and absolute. Second, object dependencies are resolved via a placement priority list generated by another LRM, ensuring a logical ordering (e.g., placing a table before placing a plate on top of that table). During placement, the LRM relies on bounding box dimensions to determine spatial relationships. However, in dense scenes, collisions may occur because the model selectively considers only a subset of previously placed objects (typically those it deems locally relevant to the current placement). This limitation stems from constraints on reasoning depth and input length, which prevent the model from exhaustively evaluating all potential spatial conflicts across the entire scene.
To mitigate these object collisions, we introduce a refinement step. After initial placement, the LRM receives detected collisions as input and revises object positions and rotations accordingly, processing each object in the same order as during initial placement. These collisions are overlaps of the objects' bounding boxes. The detected collisions comprise a complete list of all overlapping bounding boxes with a small buffer. To alleviate the complexity of the geometric reasoning task, each object is also annotated with a ``size after rotation'' attribute, which reflects the dimensions of the axis-aligned bounding box after applying the object’s rotation. We find that this precomputed property substantially improves the model’s ability to reason about spatial constraints in transformed coordinate spaces.
Importantly, not all collisions are considered undesirable. Since reasoning is performed over bounding boxes, benign overlaps may be semantically valid (e.g., a trash bin under a table, even if the table and trash bin bounding boxes intersect). Therefore, the refinement step does not indiscriminately eliminate all collisions. Instead, the LRM evaluates each collision involving the current object being refined and determines whether it is contextually appropriate, allowing for fine-grained spatial reasoning and scene-specific tolerance of overlaps. This two-stage approach helps overcome common failure modes (e.g., wrongly ordered placements, unintended occlusions) and reduces reasoning load by injecting task-relevant geometric abstractions into the LRM's input.

\section{Results}

We conduct a comprehensive evaluation of our system across three core experiments: \emph{object retrieval accuracy}, \emph{object placement precision}, and \emph{overall scene quality}. Additionally, we benchmark various state-of-the-art reasoning language models on a suite of spatial reasoning and planning tasks using the Reason-3D framework. Furthermore, we demonstrate generalization through the synthesis of complex outdoor scenes. We conduct human evaluation studies using an online tool that provides access to trained participants. More information on the human evaluation and the Elo metric calculation can be found in \appref{app:human_eval}.

\paragraph{Implementation details.}
In all experiments, we use the Gemini 2.5 model~\cite{comanici2025gemini25pushingfrontier} for object placement, object refinement, and image captioning (VLM). While we tested the pipeline on other LRMs, the results showed that Gemini 2.5 consistently yielded the best output quality. We use Gemini \texttt{embedding-004} as the embedding model.

\paragraph{Baselines.}
We compare our system, Reason-3D, against two recent scene synthesis frameworks: Holodeck~\cite{yang2024holodeck} and LayoutVLM~\cite{sun2025layoutvlm}. Holodeck is a multi-agent system that uses LLMs to generate domain-specific layout programs, which are later optimized using layout engines. LayoutVLM, on the other hand, uses a VLM to produce visually grounded spatial representations and a self-consistency decoding mechanism to enhance spatial layout planning.
We utilize the publicly released Objaverse subset provided by the LayoutVLM source code. This dataset is converted using Objathor~\cite{yang2024holodeck} for compatibility with Holodeck, while Reason-3D uses its own preprocessing. Since LayoutVLM does not support object retrieval, we supply it with objects retrieved using Reason-3D for the full scene synthesis experiments. For experiments focusing on layout design, the same set of objects is predetermined and supplied to both Reason-3D and LayoutVLM, alongside the textual prompts. In summary, we compare Reason-3D against Holodeck on the retrieval task, against LayoutVLM on layout generation, and against both systems in the full scene synthesis evaluation.

\paragraph{Full scene synthesis.}
We evaluate full scene synthesis using only a text prompt and room dimensions (see~\cref{fig:extended_comparision}) for Reason-3D, LayoutVLM~\cite{sun2025layoutvlm}, and Holodeck~\cite{yang2024holodeck}. LayoutVLM receives additional inputs, including a predefined object list and room configuration via its configuration file. In \cref{tab:win_rate_matrix}, we present the results of the human preference study using 60 participants. We find that Reason-3D produces more semantically aligned and spatially valid scenes compared to both baselines.

\begin{table}[t]
\centering
\renewcommand{\arraystretch}{0.9}
\small
\begin{tabularx}{\columnwidth}{
p{2.3cm}
>{\centering\arraybackslash}p{0.8cm}
>{\centering\arraybackslash}p{0.8cm}
>{\centering\arraybackslash}X
>{\centering\arraybackslash}p{0.8cm}
}
\toprule
\textbf{Model} &  \multicolumn{3}{c}{\textbf{Win-rates} [\%]} & \textbf{Elo} $\uparrow$ \\
\midrule
 & HD & LVLM & Reason-3D &  \\
\midrule
HD & -- & 26.9 & 4.8 & 1500 \\
LVLM & 73.1 & -- & 1.6 & 1650 \\
Reason-3D (ours) & \textbf{95.2} & \textbf{98.4} & -- & \textbf{2248} \\
\bottomrule
\end{tabularx}
\caption{Results of human evaluation comparing our approach to baselines. Win-rates show the percentage of times the row model won against the column model in head-to-head comparisons. Higher Elo scores indicate stronger performance. Results show that participants strongly prefer Reason-3D (R-3D) over both Holodeck (HD) and LayoutVLM (LVLM) in head-to-head comparisons.}
\label{tab:win_rate_matrix}
\end{table}

\begin{table}[t]
\centering
\renewcommand{\arraystretch}{0.9}
\small
\begin{tabularx}{\columnwidth}{
p{2.5cm}
>{\centering\arraybackslash}X
>{\centering\arraybackslash}X
>{\centering\arraybackslash}X
} 
\toprule
\textbf{Model} & \textbf{Top-1} & \textbf{Top-5} & \textbf{Top-10} \\ 
\midrule
Holodeck & 7\% & 8\% & 8\% \\
Reason-3D (ours) & \textbf{75\%} & \textbf{85\%} & \textbf{90\%} \\ 
\bottomrule
\end{tabularx}
\caption{Results of the retrieval experiment. Top-$k$ accuracies indicate the fraction of time the correct object was present in the top-$k$ retrieved objects.}
\label{tab:retrieval_rates}
\end{table}

\begin{table}[!t]
\centering
\renewcommand{\arraystretch}{0.9}
\small
\begin{tabularx}{\columnwidth}{p{2.5cm}
>{\centering\arraybackslash}X
>{\centering\arraybackslash}X
>{\centering\arraybackslash}X
>{\centering\arraybackslash}X
>{\centering\arraybackslash}X}
\toprule
\textbf{Model} & 1 & 2 & 3 & 4 & 5 \\
\midrule
LayoutVLM & 2.8 & 3.4 & 3.0 & 2.5 & 2.4 \\
Reason-3D (ours)  & \textbf{4.4} & \textbf{3.9} & \textbf{4.4} & \textbf{4.1} & \textbf{4.3} \\
\bottomrule
\end{tabularx}
\caption{Results of the human evaluation on object placement. Each column indicates the instruction complexity (see~\cref{fig:evolution_classroom}). All results are below 0.2 standard deviation and on a scale from 1 (bad) to 5 (excellent). Overall, we observe that participants prefer Reason-3D over previous work, and that as the instruction complexity increases, Reason-3D can place objects more reasonably than previous work.}
\label{fig:placement_evaluation}
\end{table}

\begin{figure}[t]
  \centering
  \includegraphics[width=0.98\columnwidth]{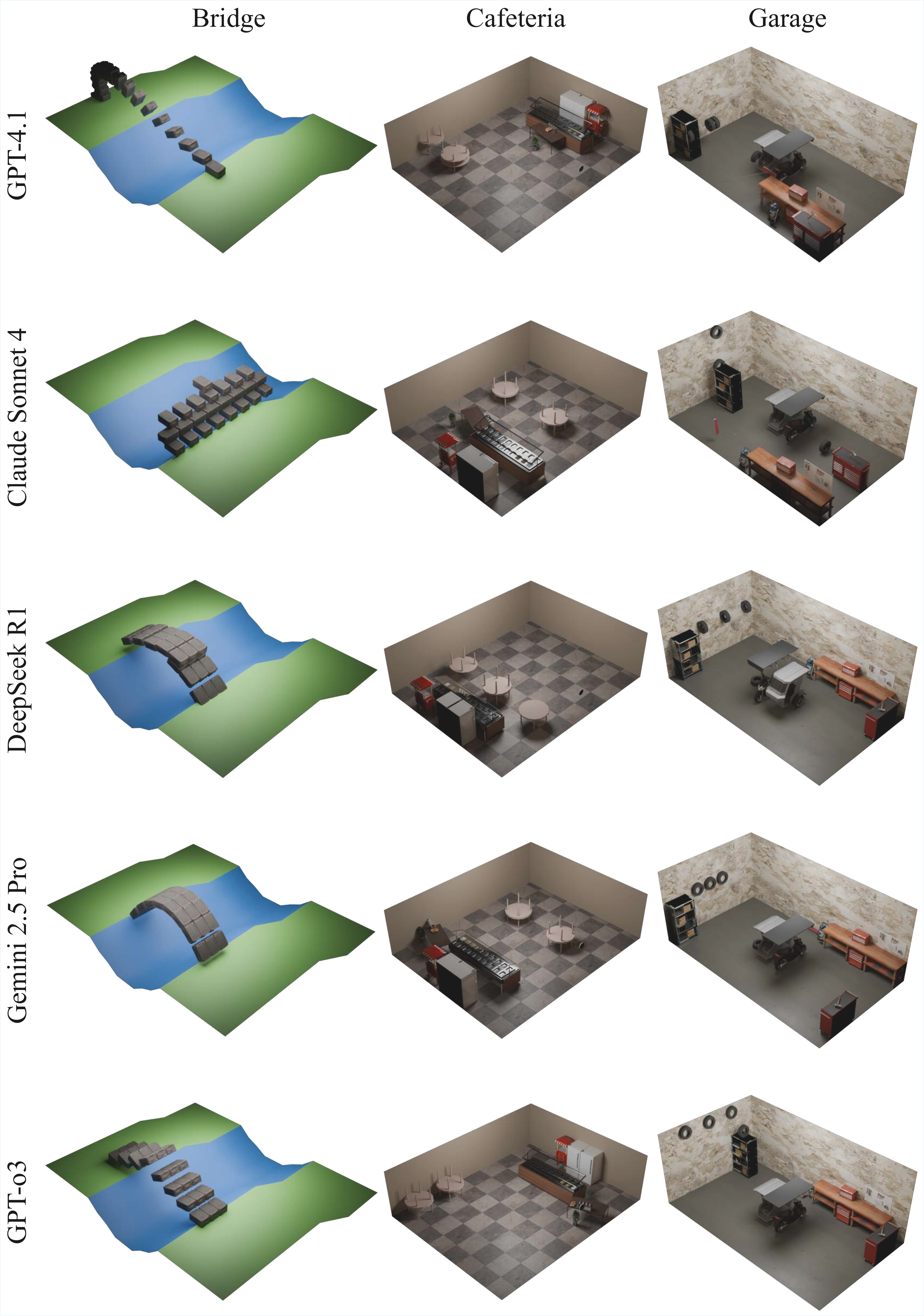}
  \caption{We benchmark various LRMs on three scenes. We find that Gemini 2.5 Pro achieves the best overall performance on spatial reasoning tasks. The bridge-building task also showcases the ability of object composition (building a bridge out of individual stone objects).}
  \label{fig:reasoning_benchmark}
\end{figure}

\begin{table*}[!th]
\centering
\small
\renewcommand{\arraystretch}{1}
\begin{tabularx}{\linewidth}{p{2.2cm} 
>{\centering\arraybackslash}p{1.2cm}
>{\centering\arraybackslash}X
>{\centering\arraybackslash}X
>{\centering\arraybackslash}X
>{\centering\arraybackslash}p{2.2cm}
>{\centering\arraybackslash}p{0.8cm}
>{\centering\arraybackslash}p{0.8cm}
>{\centering\arraybackslash}p{0.9cm}}
\toprule
\textbf{Models} & &  \multicolumn{3}{c}{\textbf{Win-rates} [\%]}  & & \textbf{Elo} $\uparrow$ & \multicolumn{2}{c}{\textbf{Tokens}}\\
\midrule
  & GPT-4.1 & Sonnet 4 & DeepSeek-R1 & GPT-o3 & Gemini 2.5 Pro & & Input & Output \\
\midrule
GPT-4.1      & --   & 38.5 & 18.2 & 0.0 & 10.0 & 1500 & 41,424 & \phantom{000,}971\\
Sonnet 4     & 61.5 & --   & 14.3 & 11.1 & 8.3 & 1566 & 52,225 & \phantom{00}1,698\\
DeepSeek-R1  & 81.8 & 85.7 & --   & 40.0 & 7.7 & 1809 & 47,426 & 114,506\\
GPT-o3       & 100.0& 88.9 & 60.0 & --  & 29.4 & 1938 & 39,770 & \phantom{0}35,249\\
Gemini 2.5 Pro & \textbf{90.0} & \textbf{91.7} & \textbf{92.3} & \textbf{70.6} & -- & \textbf{2091} & 48,470 & \phantom{0}29,312\\
\bottomrule
\end{tabularx}
\caption{Results of human evaluation on LRM benchmark. Each cell shows the percentage of times the row model won against the column model in head-to-head comparisons. The token statistics for different models were computed on a per-scene basis, assuming an average of 14 objects per scene. While the calculation and accounting of ``reasoning tokens'' vary across models, they are generally reflected in the output token count. GPT-4.1, which does not support multi-step reasoning, produces the fewest output tokens. In contrast, DeepSeek generates the highest number of output tokens, indicating more extensive reasoning during scene construction. We find that Gemini 2.5 Pro performed the best overall, achieving the highest Elo score.}
\label{tab:benchmark_win_rate_matrix}
\end{table*}

\paragraph{Retrieval benchmark.}
To evaluate retrieval accuracy, we assess whether objects described in a scene prompt can be correctly retrieved from the object database. For controlled testing, we generate scene descriptions from a predefined list of target objects using an LLM, with each object accompanied by a rich semantic description. During retrieval, each target object is compared to all entries in the database via cosine similarity of embedding vectors. The results yield a ranked list of the top 10 most similar objects. We report top-1, top-5, and top-10 accuracy, where top-$k$ indicates the proportion of cases in which the correct object is ranked within the first $k$ retrieved candidates. This evaluation is averaged across multiple runs with a 12-object test set. As observed in \cref{tab:retrieval_rates}, we find that Reason-3D outperforms Holodeck in retrieval accuracy.

\paragraph{Object placement benchmark.}
We compare object placement quality between Reason-3D and LayoutVLM. Starting from a simple prompt and a minimal set of objects, we incrementally increase scene complexity by adding objects with layout constraints. We present a qualitative example of this experiment in \cref{fig:evolution_classroom}, illustrating a five-step iteration comparison. In \cref{fig:placement_evaluation}, we present the results of a human evaluation study with 43 participants asked to rate the quality of the scene and its adherence to the provided constraints. We find that Reason-3D consistently produces more spatially coherent placements as the instruction complexity increases.
Reason-3D can infer object orientation from bounding box dimensions, successfully aligning objects such as a periodic table, an area in which previous methods like LayoutVLM offer no necessary rotational support across all axes. More details and examples on this can be found in~\appref{app:placement_experiment}.

\begin{figure}[t]
  \centering
  \includegraphics[width=0.8\linewidth]{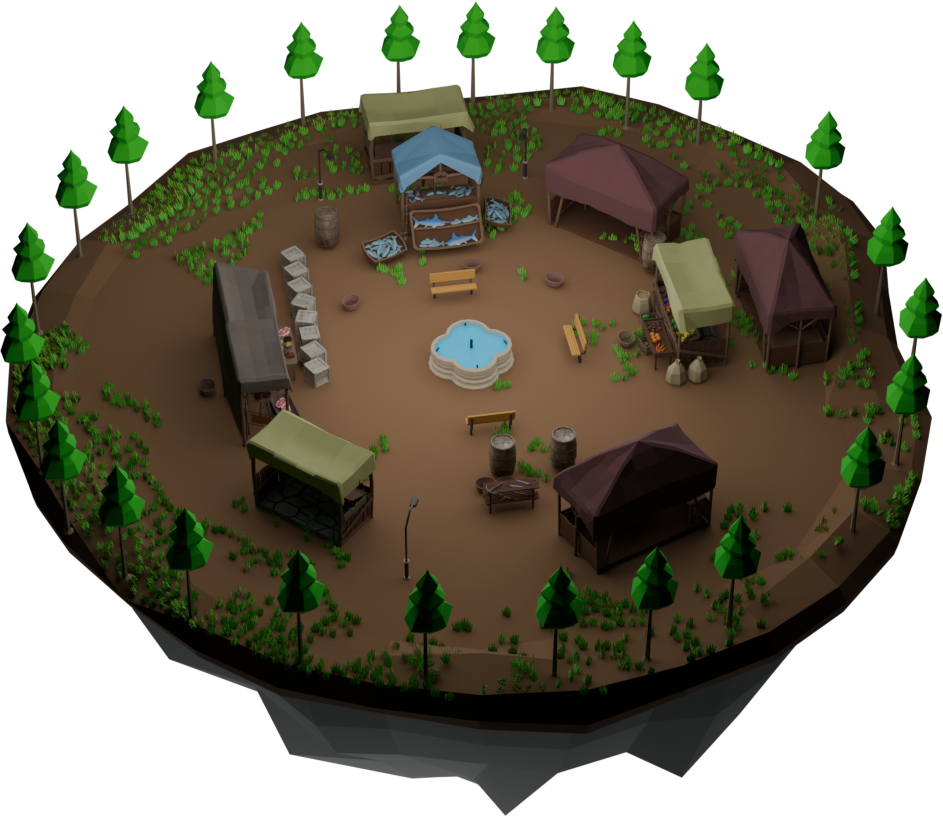}
  \caption{Qualitative example of an outdoor scene generated with Reason-3D, containing 70 objects. Instruction: \textit{A big circular floating island. On top are 25 trees surrounding the island on its edge. In the middle, on top of the island, is a water fountain. Around this fountain unfolds a lively medieval marketplace, with a fish stand, a meat stand, a weapon stand, and a vegetable stand.}}
  \label{fig:island}
\end{figure}

\paragraph{Large Reasoning Model benchmark.}
To assess spatial reasoning capabilities in isolation, we benchmark the following LRMs: Gemini 2.5 Pro~\cite{comanici2025gemini25pushingfrontier}, GPT-o3~\cite{openai2025o3_o4mini}, Claude Sonnet 4~\cite{anthropic2025claude4}, DeepSeek-R1~\cite{deepseek2025R1}. Additionally, we use GPT-4.1~\cite{openai2025gpt4_1} as a baseline due to its lack of test-time compute (i.e., no explicit multi-step reasoning). The evaluation covers three dimensions of spatial reasoning. First, we test the ability to reason in a transformed coordinate system, requiring the model to apply correct geometric transformations. Second, we assess whether the model can infer object orientation based on the dimensions of its axis-aligned bounding box. Finally, we combine both challenges in a complex spatial planning task, requiring integrated multi-object reasoning. We test the LRMs on three scenes (more information can be found in \appref{app:benchmark_scenes}). We conducted a human evaluation with 40 participants, where each participant rated a scene synthesized by two models in a head-to-head comparison. The results are presented in \cref{tab:benchmark_win_rate_matrix}, with comparisons illustrated in \cref{fig:reasoning_benchmark}. Furthermore, \cref{tab:benchmark_win_rate_matrix} summarizes the average token usage for each model, providing insight into the computational cost associated with different levels of reasoning.

\paragraph{Outdoor scenes.}
We demonstrate Reason-3D’s ability to generalize beyond indoor environments by synthesizing several outdoor scenes. \cref{fig:island} presents an open-world example composed of 70 objects, illustrating Reason-3D's capability to handle large-scale layouts and diverse spatial contexts. This highlights the flexibility of the system and its applicability across a wide range of scene types. Additional examples in \appref{app:additional_outdoor_scenes}.

\paragraph{Limitations.}
Our framework assumes that the object database contains objects that are pre-aligned upright. Misalignment involving other rotational degrees of freedom cannot be resolved currently, as they would require additional processing. Additionally, object scales must be approximately correct between objects. Furthermore, our framework does not generate structural elements such as floors or walls. More generally, Reason-3D depends on LRMs, implicitly inheriting all of their limitations.

\section{Conclusion}
We presented Reason-3D, a text-driven framework for 3D scene synthesis that leverages the multi-step reasoning capabilities of LRMs. Our results demonstrate that, without requiring task-specific fine-tuning, these off-the-shelf models can retrieve and place objects more desirably than previous baselines. Given the rapid evolution of these models, we anticipate their continued integration and refinement in future research.

{\small
\bibliography{references}
}

\ifisextended
    \appendix
    \onecolumn
\section{Model Instructions}
\label{app:prompts}

In the following we describe the various prompts used in our pipeline. For the vision-language model we use the following prompts:
\begin{promptbox}{Captioning prompt for object orientation preprocessing}
\small
\ttfamily
You are given four images of the same object from different angles. The name of the object is \{name\}. Your job is to tell me in what image the the object is shown from the front and you can clearly see and identify it. If the object is radially symmetrical regarding their primary structure, pick an image where the object looks natural. In order to tell me which image you chose, give me as output a number between 1 and 4, which serves as the index for the list of images I provide.
\end{promptbox}
\begin{promptbox}{Captioning prompt for extracting object properties}
\small
\ttfamily
The two images show the same object from different angles/perspectives. Please provide a detailed structured description of this object divided into these three categories: (1) Physical properties (size, shape, color, material, parts/components), (2) Functional properties (purpose, how it's used, what it does), and (3) Contextual properties (where it might be found, what settings it belongs in). You may use the name of the object as a hint, if it is helpful: \{name\}.
\end{promptbox}
\noindent An example output, including the two rendered input images, is shown in \cref{fig:description_example}. To extract a list of \texttt{num\_obj} objects from a scene description, we prompt the LLM as follows:
\begin{promptbox}{Prompt to list possible objects \& their attributes from a scene description}
\small
\ttfamily
Given the following scene description: \{scene\_description\}. \\ 
Please follow these steps: \\
Step 1: Identify \{num\_obj\} objects in the scene. If you think there are more objects focus on the most important ones. Try to be a bit broad and don't pick objects that that are part of any other objects that you picked. The objects can't be entire rooms or places. The objects also cannot be walls, floors, ceilings, doors or anything else that has to do with the
frame of the room. It also can't be the ground, water, sky or the sun. \\
Step 2: Without the context of the scene, describe each object, focusing on its physical properties, functional properties, and contextual properties. Be detailed. \\
Step 3: Tell me how many duplicates of the object are required.
\end{promptbox}
\noindent The LLM voting prompt in the retrieval step is shown below.
\begin{promptbox}{Prompt to pick the closest match or exclude the object from the scene}
\small
\ttfamily
You are given a description of a target object and a list of five objects with their descriptions. Your task is to tell me which one of the five objects in the list matches the description of the target object best. Do this by giving me a number between 1 and 5, which serves as an index of the list. If you think that no object in the list can be can be used as a substitution for the target object, please output a 0. 
\end{promptbox}
\noindent Then, given a list of object and a scene description, we extract the implicit and explicit constrains using the following:
\begin{promptbox}{Prompt to extract constraints}
\small
\ttfamily
You are given a scene description and a list of objects that are part of that scene. Your task is to give me positional and rotational constraints regarding the objects. If the scene description is vague and doesn't contain any positional and rotational information, I need you to add this information to create a scene that makes sense. Especially important is where the objects are placed upon or if they are against a wall and if there should be space between them. There should be a constraint for each object about this. E.g. the object is standing on the ground. The object is on the table. The object is against the north wall. etc. You are only allowed to use objects from the list to write constraints. Refrain from absolute measurements like 6 feet and so on. Only output the constraints. Under no circumstances should you contradict constraints from the scene description. Start by extracting the constraints that are already part of the scene description and then add the other constraints.
\end{promptbox}
\noindent We obtain the optimal object placement order using the following prompt.
\begin{promptbox}{Prompt to get the placement order}
\small
\ttfamily
You are given a list of constraints on objects about their placements and rotations. You also get the list containing all the objects. Your goal is to sort the list such that placing the objects one by one is easiest. For example, if you have a constraint: The cup is on the table. You want to place the table before the cup. IMPORTANT: Under no circumstances should you add or remove any objects from the list.
\end{promptbox}
\noindent The objects are then placed using the following:
\begin{promptbox}{Prompt for the initial placement of the objects}
\small
\ttfamily
System instructions \\
You are an expert AI assistant specializing in 3D object placement for the Unity game engine. Your task is to determine the correct position and rotation for a new object based on a scene description and a list of existing objects. This task requires a lot of complex reasoning and some math.
Prompt
CRITICAL CONTEXT: UNITY'S 3D SPACE: \\
You MUST adhere to these rules at all times. All calculations and outputs must conform to Unity's coordinate system. \\ \\
Coordinate System: Left-Hand System (LHS). \\
+X axis: Right \\
+Y axis: Up \\
+Z axis: Forward \\
Default Object Orientation: An object with zero rotation (0, 0, 0) faces the positive Z-axis (0, 0, 1) (+Z). \\
Rotation Rule: Rotations are Euler angles (X, Y, Z) in degrees. \\
A positive rotation around the Y-axis rotates an object from the Forward direction (Z+) towards the Right direction (X+). \\
A negative rotation around the Y-axis rotates an object from the Forward direction (Z+) towards the Left direction (X-). \\
The floor is at Y = 0. \\ \\
YOUR TASK: \\
1. Analyze the Scene: Read the scene\_description and the already\_placed\_objects list. \\
2. Determine Placement: Calculate the position and rotation for the new object, \{object\_name\}, which has a bounding box size of \{object\_size\}. \\Think where this object should be naturally placed and which way it should be facing, if there are no specific constraints on that matter.. \\
3. Adhere to Constraints: Ensure the placement satisfies all layout rules from the scene\_description and the constraints list. \\
4. Output JSON: Generate a single, clean JSON object with the final position and rotation. \\ \\
INPUT DATA: \\
Scene Description: \{scene\_description\} \\
Constraints: \{constraints\} \\
New Object to Place: \\
name: \{object\_name\}
size: \{object\_size\}
Already Placed Objects: \{placed\_objects\} \\
Each object in the list has a name, center, size, size\_after\_rotation and rotation. \\
The size\_after\_rotation field is the original size of the bounding box with the applied rotation. This is to save you the trouble of doing the math yourself.
\end{promptbox}

\begin{figure*}[t]
  \centering
  \includegraphics[width=\textwidth]{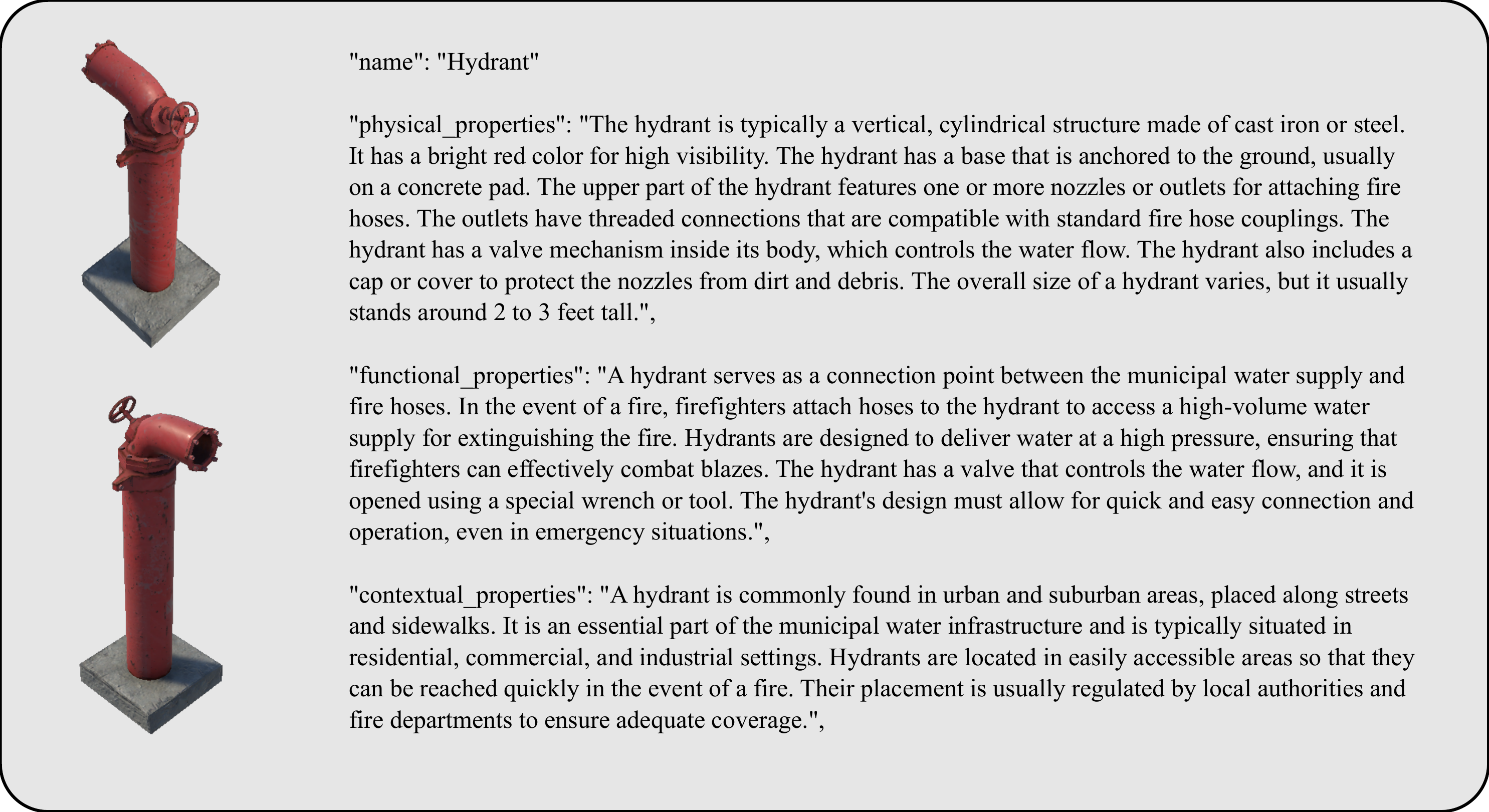}
  \caption{This is an example of the renders captured of an object and then annotated by the VLM.}
  \label{fig:description_example}
\end{figure*}






\section{Human Evaluation}
\label{app:human_eval}

\paragraph{Details on human evaluation.} We use Mabyduck\footnote{\url{https://www.mabyduck.com/}} to conduct the human evaluation. Mabyduck conducts pre-screening tests and keeps the best performing participants in a pool from which our raters are drawn. These participants are paid above average rates compared to Prolific or other crowd-sourcing platforms. \cref{fig:showcase_interface_human_comparison,fig:placement_interface_human_comparison} demonstrate the interface for the human raters for the overall scene synthesis comparison, and the object placement benchmark, respectively.

\paragraph{Elo rating calculation.} To compute relative model performance rankings, we applied the Bradley-Terry model using Maximum Likelihood Estimation (MLE) to the pairwise comparison data from our human evaluation study. First, we extracted all pairwise comparisons from the evaluation dataset, where each comparison represented a head-to-head judgment between two models on the same stimulus. We constructed win matrices recording the number of times each model defeated every other model in direct comparisons. The Bradley-Terry model parameters ($\pi_i$ values representing each model's inherent strength) were estimated iteratively using MLE, where each model's strength is updated as $$\pi_i = \frac{\sum w_{ij}}{\sum \frac{n_{ij}}{\pi_i + \pi_j}},$$ with $w_{ij}$ representing wins of model $i$ over model $j$ and $n_{ij}$ representing total comparisons between the models. The algorithm converged after 184 iterations with a tolerance of 1e-8. Finally, we converted the Bradley-Terry strength parameters to traditional Elo ratings using the formula: $$\mathrm{Elo}_i = 400 \times \log_{10}\left(\frac{\pi_i}{\pi_{\text{baseline}}}\right) + 1500,$$ where the baseline model (Holodeck) was assigned the standard starting Elo rating of 1500.

\paragraph{Statistical Significance.} For the overall retrieval and placement performance, a human evaluation with 60 raters and 333 pairwise comparisons revealed statistically significant differences between all three models ($\chi^2 = 126.01$, $p < 0.001$). All pairwise comparisons
were statistically significant using Fisher's exact tests (all $p < 0.001$),
even after Bonferroni correction for multiple comparisons. For the reasoning benchmark evaluation with 120 pairwise comparisons across 5 models, we found significant performance differences ($\chi^2 = 29.77$, $p < 0.001$). Head-to-head Fisher's exact tests confirmed significant differences between top-performing models, even after Bonferroni correction for multiple comparisons. In the placement evaluation with 639 ratings, Reason-3D
achieved significantly higher ratings than LayoutVLM ($M = 4.21$ vs.\ $M = 2.80$, $t(637)=16.80$, $p<.001$, Cohen's $d = 1.33$). The 1.40-point difference represents a large effect size.

\section{Additional Examples}

\subsection{Placement experiment}
\label{app:placement_experiment}

We show further results from the placement experiment against LayoutVLM. \cref{fig:evolution_waiting_room} shows the evolution of prompting the model with additional objects and constraints. This prompt in particular is testing the model on consistency with prompted multiple times with the same layout constraints. It also tests the ability of making order in the layout where required, as with the chairs, that are to be placed in a row. \cref{fig:evolution_chemistry} shows another example of such an experiment. This scene tests the overall flexibility of the model in terms of object transformations, specifically rotations, testing the ability to rotate objects precisely as requested. LRMs have the ability to do math when reasoning, allowing them to calculate correct rotations when necessary. The whiteboard, which is rotated exactly as the prompt requested, demonstrates this behavior. Additionally, these tests examine the following: (1) Can the model notice when an object is initially not correctly aligned, and (2) can the model then rotate objects around all necessary axes to rectify this misalignment. The correctly placed and rotated \textit{periodic table of elements} object demonstrates this ability of Reason-3D.

\begin{figure*}[!t]
  \centering
  \includegraphics[width=\textwidth]{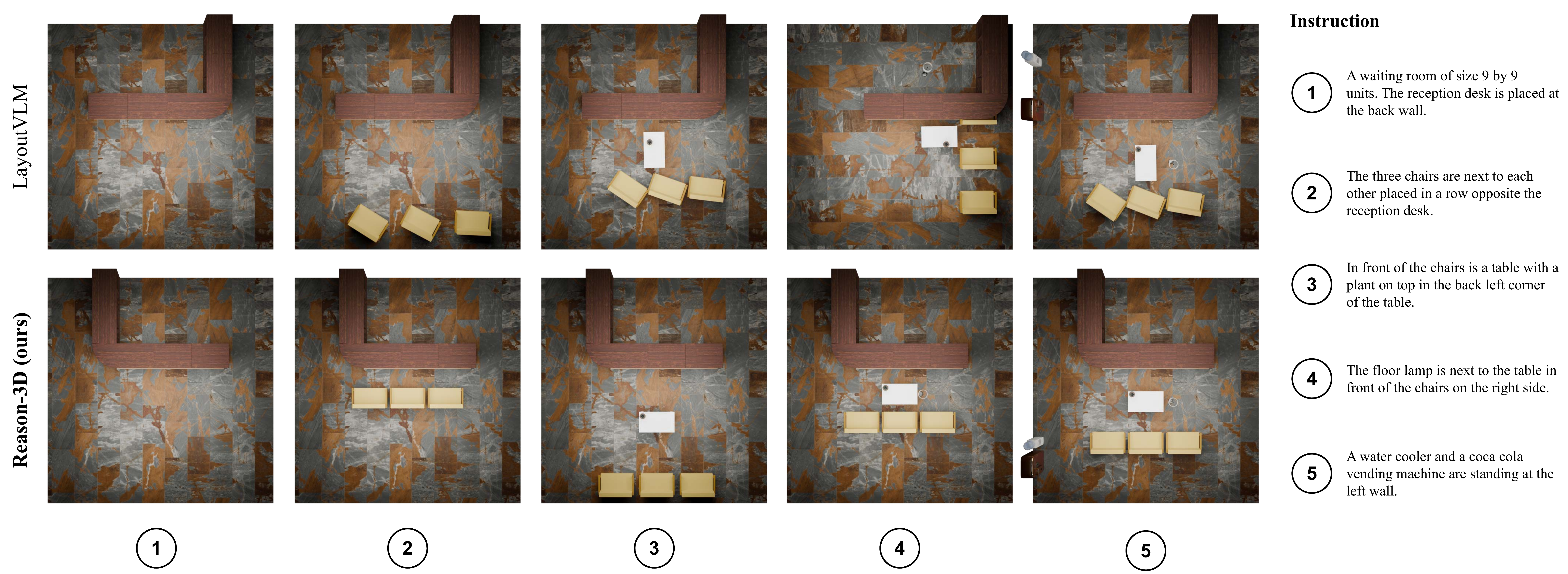}
  \caption{Qualitative comparison of object placement performance when instruction complexity is increased. Every scene is generated from scratch with the entire instruction given up to and including its circled number.}
  \label{fig:evolution_waiting_room}
\end{figure*}

\begin{figure*}[!t]
  \centering
  \includegraphics[width=\textwidth]{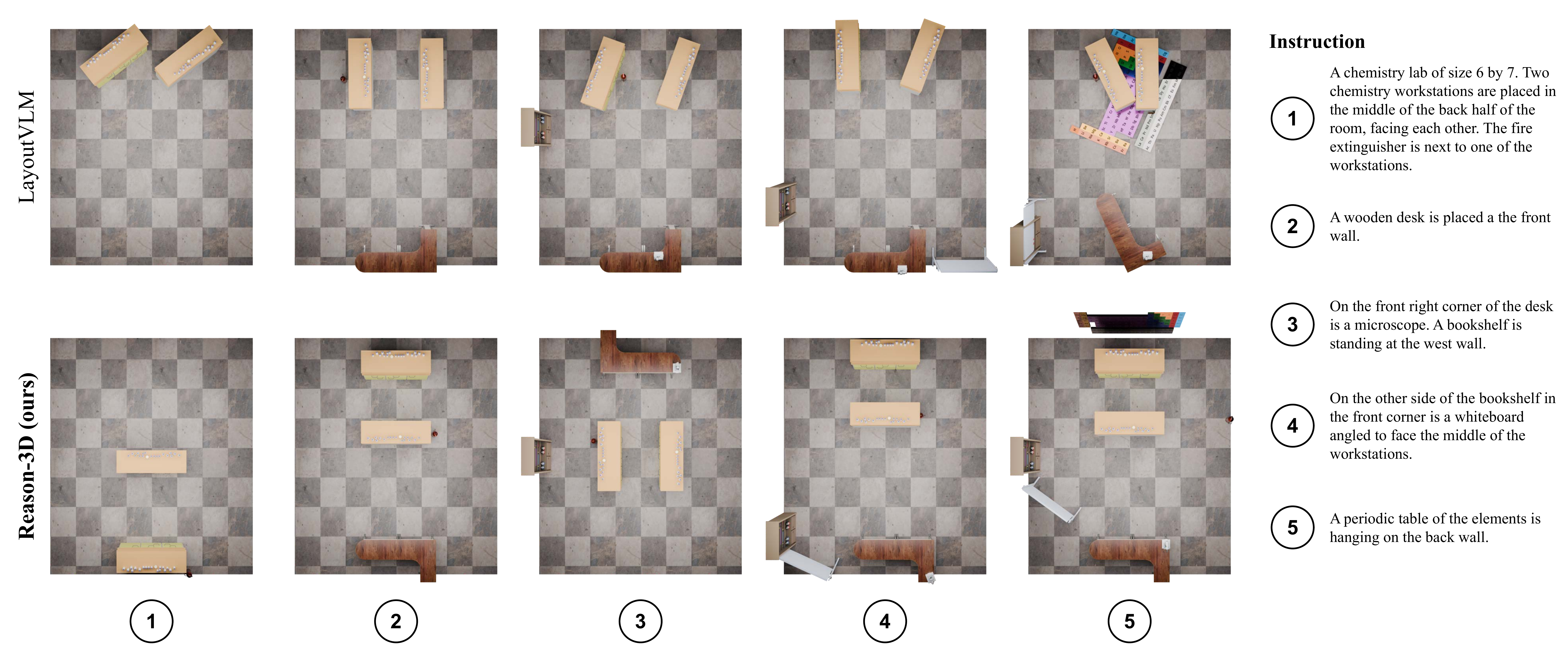}
  \caption{Qualitative comparison of object placement performance when instruction complexity is increased. Every scene is generated from scratch with the entire instruction given up to and including its circled number.}
  \label{fig:evolution_chemistry}
\end{figure*}

\subsection{Additional outdoor scenes}
\label{app:additional_outdoor_scenes}
In this section, we present additional outdoor scene generations based on a range of prompt types, from vague or ambiguous to highly detailed instructions. \cref{fig:gas_station} illustrates a scene generated from a prompt that specifies object retrieval with minimal layout constraints, allowing the model to infer spatial relationships freely. \cref{fig:park} demonstrates the model’s ability to interpret and execute a loosely defined prompt involving general object inclusion and placement. In contrast, \cref{fig:patio} depicts a scene generated from a prompt with precise specifications for both object identities and their spatial configurations. These examples highlight the model’s versatility: it can follow detailed instructions accurately while also generating coherent and plausible layouts in unbounded 3D environments, when guidance is minimal.

\begin{figure*}[t]
  \centering
  \begin{subfigure}[b]{0.34\linewidth}
    \includegraphics[width=\linewidth]{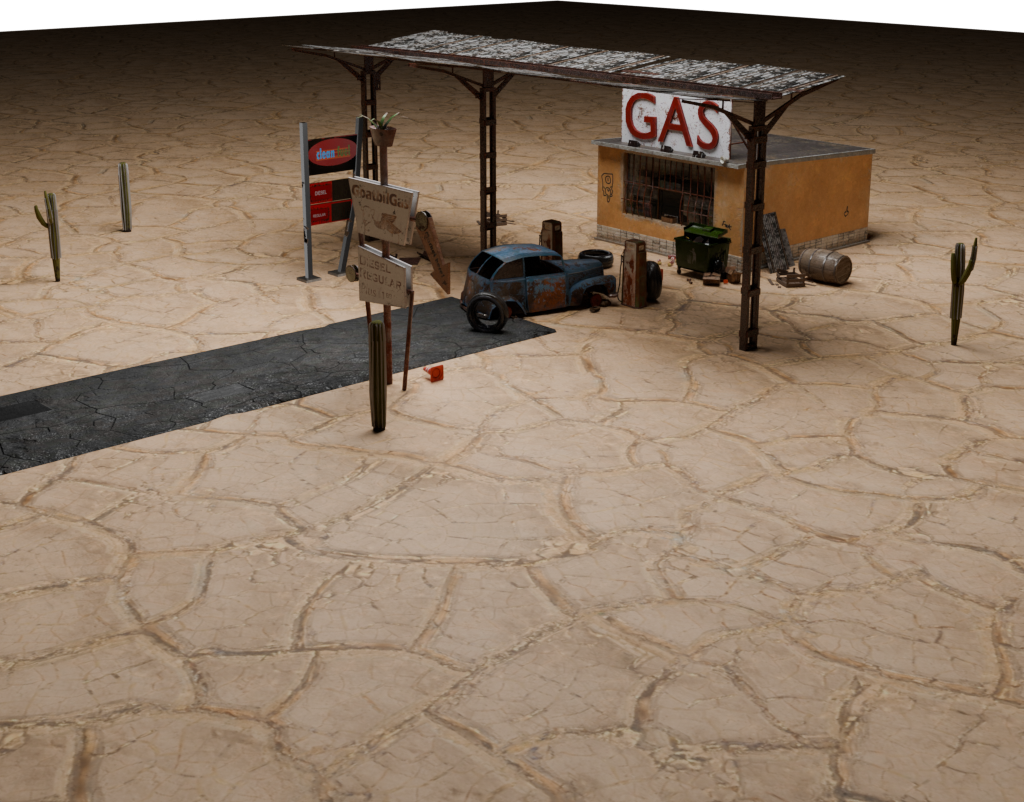}
    \caption{Instruction: \textit{An abandoned gas station in the desert. There should be a station canopy and an operator shack. Add desert plants.}}
    \label{fig:gas_station}
  \end{subfigure}
  \hfill
  \begin{subfigure}[b]{0.27\linewidth}
    \includegraphics[width=\linewidth]{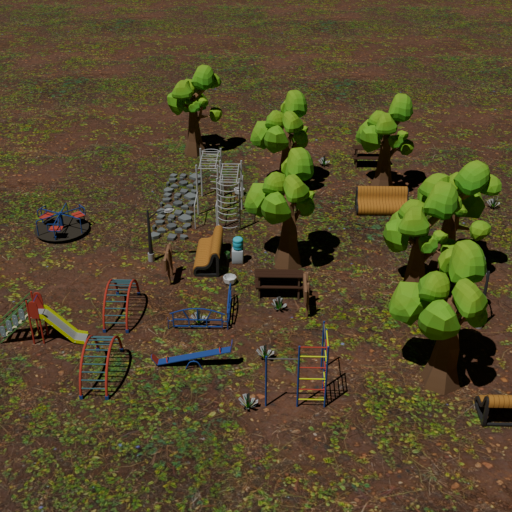}
    \caption{Instruction: \textit{A big outdoor park in nature, with a section for a kids playground.}}
    \label{fig:park}
  \end{subfigure}
  \hfill
  \begin{subfigure}[b]{0.34\linewidth}
    \includegraphics[width=\linewidth]{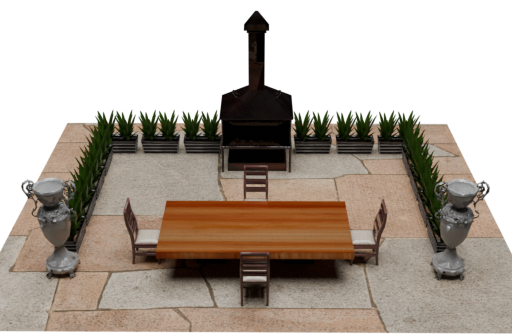} 
    \caption{Instruction: \textit{The scene shows a cozy outdoor dining setup. The plants are arranged to form a U-shape with the oven at the base of the U.}}
    \label{fig:patio}
  \end{subfigure}
  \caption{Examples of outdoor scenes generated by Reason-3D. Each image corresponds to a textual instruction (see subcaptions).}
  \label{fig:outdoor_examples}
\end{figure*}

\subsection{Additional metrics}
Similar to \citet{sun2025layoutvlm}, we measure physical and semantical coherence using separate metrics. The most relevant metric that measures plausible physical placement is the Collision Free score, measuring how many objects are colliding in the final scene. We split semantical coherence into positional and rotational accuracy and query GPT-5 to rate them on a scale of 0 to 100 based on the scene description and a rendering of the final scene. This is the exact same approach as \citet{sun2025layoutvlm}, utilizing the same prompts.

\begin{table}[t]
\centering
\begin{tabularx}{\columnwidth}{p{2.5cm}>{\centering\arraybackslash}X>{\centering\arraybackslash}X>{\centering\arraybackslash}X} 
\toprule
\textbf{Model} & \textbf{Pos. Coherence} & \textbf{Rot. Coherence} & \textbf{Collision Free} \\ 
\midrule
Holodeck & 47 & 57 & 87.5\% \\
LayoutVLM & 76.5 & 74.5 & 71.4\% \\
Reason-3D (ours) & \textbf{81} & \textbf{77} & \textbf{95.9\%} \\ 
\bottomrule
\end{tabularx}
\caption{Results of the VLM scoring the scenes on positional and rotational coherence to the description, and a score for object collision freeness.}
\label{tab:VLM_scores}
\end{table}

\section{Scenes for Reasoning Ablation}
\label{app:benchmark_scenes}
The reasoning scenes are carefully designed to test different models on different reasoning tasks. The objects are chosen beforehand and passed to the model instead of the traditional retrieval step. The rest of the pipeline remains the same.
For the three scenes used to evaluate reasoning we use the following prompts:
\begin{promptbox}{Prompt with the criteria for the bridge scene}
\small
\ttfamily
1: There should be a bridge. 2: The bridge should be smooth, with no steps. 3: The bridge looks good.
\end{promptbox}
\begin{promptbox}{Prompt with the criteria for the cafeteria scene}
\small
\ttfamily
1: There should be two pairs of tables that are correctly stacked on top of each other. 2: The wooden table needs to be an extension of the buffet with an angle. 3: The bread basket and metal dispenser need to be fully on that wooden table. 4: The scene looks good.
\end{promptbox}
\begin{promptbox}{Prompt with the criteria for the garage scene}
\small
\ttfamily
1: There should be three tires hanging on the wall correctly. 2: There should be exactly one tire lying on the ground correctly. 3: The hammer should be lying on the toolbox correctly. 4: The scene looks good.
\end{promptbox}
\noindent These scenes are testing important spatial reasoning capabilities, that vary in difficulty. The cafeteria scene includes four tables, where the LRM has to flip a table on its head, requiring a rotation around the appropriate axis. Additionally, it is asked to place items on a rotated table. Working with axis aligned bounding boxes makes this task essentially a placement task in a rotated coordinate system with more advanced reasoning steps. The garage scene tests the LRMs ability to correctly infer an objects orientation from its bounding box dimensions. The hammer object is not lying on its side initially and neither is the tire. The LRM needs to reason about initial alignment on its own. Another example can be seen in \cref{fig:evolution_chemistry}. The bridge building task combines all above mentioned tasks into one and adds an additional planning step, with each placed object. (1) The LRM needs to make a plan, how to use 24 bricks to build a bridge across 8 units, while knowing the dimensions of a single brick. (2) It needs to know what axis to rotate for the desired result. (3) Working with rotated bricks is the same problem task as mentioned in the cafeteria scene. Not all models excelled at the same reasoning tasks. Furthermore, for the evaluation of Claude Sonnet 4, we do not use extended thinking, because this disables the necessary tool use.

\begin{figure}[t]
  \centering
  \begin{subfigure}{0.685\columnwidth}
    \centering
    \includegraphics[width=\linewidth]{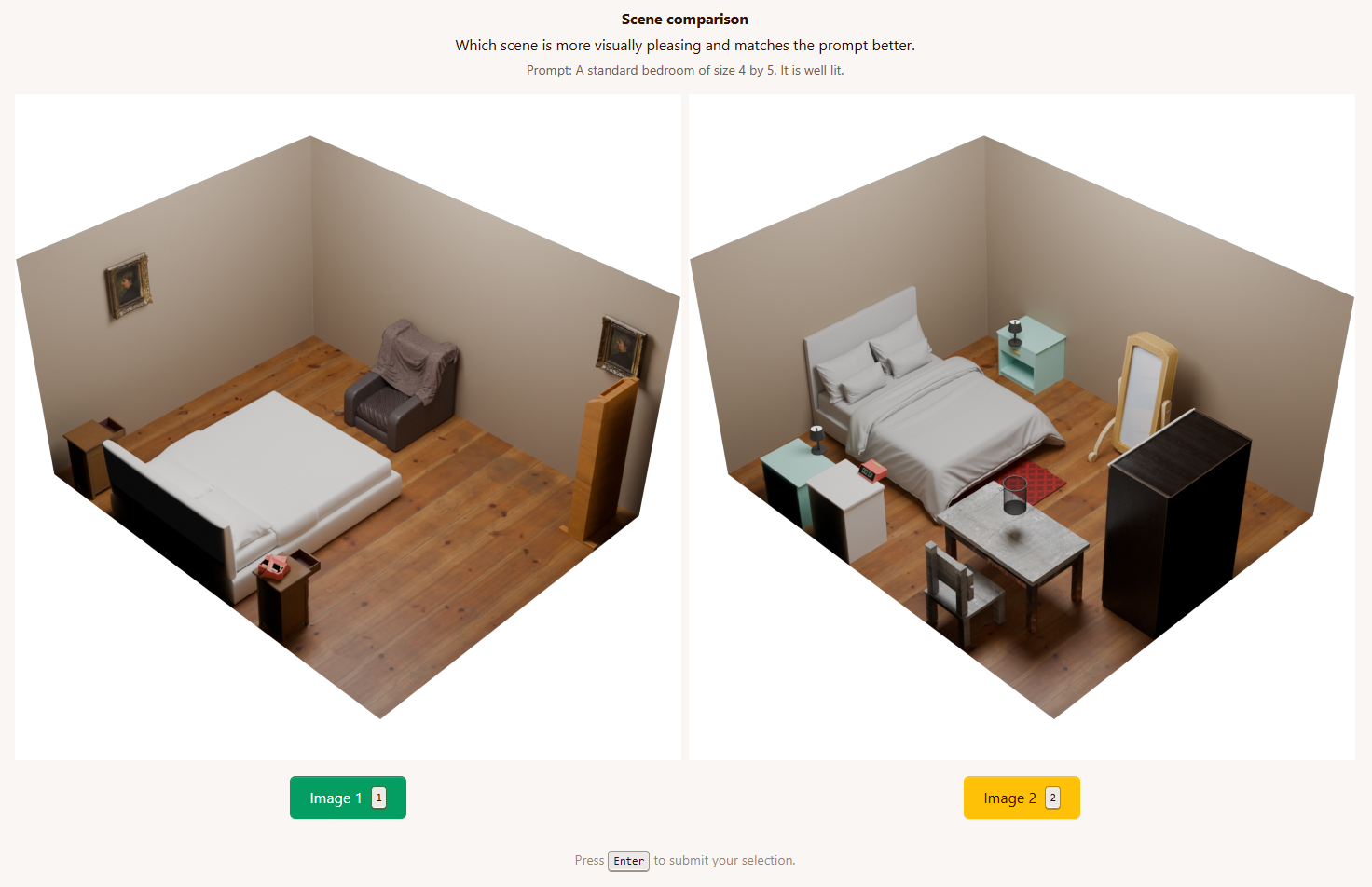}
    \caption{General scene synthesis comparison.}
    \label{fig:showcase_interface_human_comparison}
  \end{subfigure}%
  \hfill
  \begin{subfigure}{0.305\columnwidth}
    \centering
    \includegraphics[width=\linewidth]{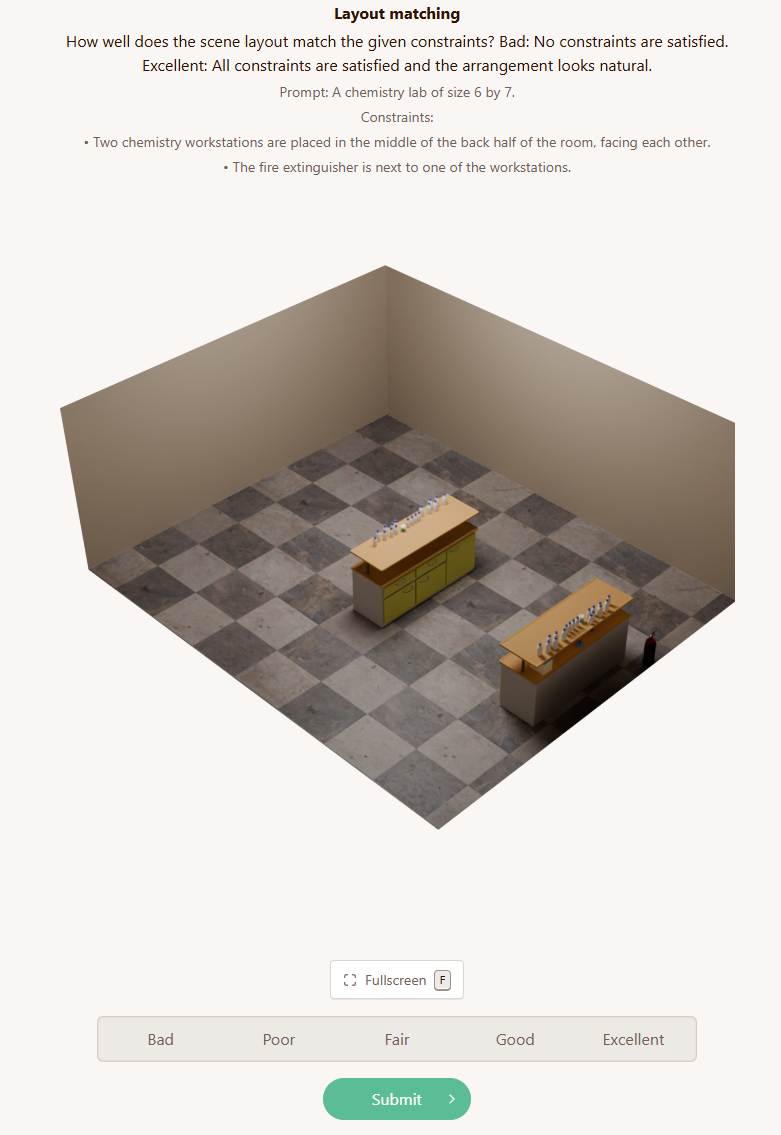}
    \caption{Object placement comparison.}
    \label{fig:placement_interface_human_comparison}
  \end{subfigure}
  \caption{Mabyduck interface used for human evaluation.}
  \label{fig:interface_comparison}
\end{figure}
\else
\fi

\end{document}